\documentclass{article}

\usepackage[preprint]{neurips_2025}

\usepackage[utf8]{inputenc} 
\usepackage[T1]{fontenc}    
\usepackage{hyperref}       
\usepackage{url}            
\usepackage{booktabs}       
\usepackage{amsfonts}       
\usepackage{nicefrac}       
\usepackage{microtype}      
\usepackage{amsmath}
\usepackage{multirow}
\usepackage{graphicx}
\usepackage{subfig}
\usepackage{amssymb}
\usepackage{pifont}
\usepackage{booktabs}
\usepackage[table,xcdraw]{xcolor}
\usepackage{comment}

\title{Unveiling the Potential of Vision-Language-Action Models with Open-Ended Multimodal Instructions}

\author{%
  Wei Zhao$^{1}$
  \quad
  Gongsheng Li$^{2}$
  \quad
  Zhefei Gong$^{1}$
  \quad
  Pengxiang Ding$^{1,2}$ \\
  \quad
  \textbf{Han Zhao}$^{1,2}$
  \quad
  \textbf{Donglin Wang}$^{1}$\thanks{Corresponding Author.} \\
  $^{1}$Westlake University\quad $^{2}$Zhejiang University
}

\begin{document}

\maketitle

\begin{abstract}
  Vision-Language-Action (VLA) models have recently become highly prominent in the field of robotics. Leveraging vision-language foundation models trained on large-scale internet data, the VLA model can generate robotic actions directly from visual observations and human instructions through a single end-to-end neural network. Despite their effectiveness, current VLA models usually accept only one form of human prompting, language instructions, which may constrain their applicability in open-ended human-robot interactions. For example, a user might expect the robot to retrieve an object shown in an image, follow an instruction written on the whiteboard, or imitate a behavior demonstrated in a video, rather than relying solely on language-based descriptions. To address this gap, we introduce OE-VLA, which explores the potential of VLA models for open-ended multimodal instructions. Extensive results demonstrate that our OE-VLA not only achieves comparable performance to traditional VLA models with linguistic input but also delivers impressive results across four additional categories of open-ended tasks. The proposed methodology could significantly expand the applications of VLA models across various everyday scenarios and facilitate human-robot interaction.
\end{abstract}

\section{Introduction}
\label{sec:1}
Large language models (LLMs), such as ChatGPT\citep{ouyang2022traininglanguagemodelsfollow}, LLaMA\citep{touvron2023llamaopenefficientfoundation}, and Gemini\citep{team_gemini_2024}, have achieved significant success in various domains and greatly eased people's daily lives. In particular, researchers have also conducted studies on multimodal large language models (MLLMs), such as vision language models (VLMs)\citep{liu_visual_2023, li_blip-2_2023, awadalla2023openflamingo, karamcheti_prismatic_2024, zhang2024llamaadapter}, to enable LLMs to understand human intentions using both linguistic and visual modalities. The aforementioned versatile models provide a robust foundation for the advancement of artificial general intelligence (AGI) within cyberspace.

As these techniques undergo continuous improvement, a critical question emerges: How can we develop similar models capable of interacting with the physical world? One possible approach is the vision-language-action (VLA) model proposed in RT-2\citep{zitkovich_rt-2_2023} for robot manipulation. This approach leverages a pre-trained VLM as the base model and fine-tunes it with carefully curated robotic data, aiming to fully harness the knowledge acquired by the VLM. Specifically, numerous studies have been conducted to investigate the potential of the VLA model. RT-2\citep{zitkovich_rt-2_2023}, RT-X\citep{collaboration_open_2023}, and OpenVLA\citep{kim_openvla_2024} demonstrate that, compared to traditional policies\citep{brohan_rt-1_2023,lynch_language_2021}, these end-to-end models can achieve superior generalization across diverse language commands and environments. Building on this starting point, numerous studies have been conducted to refine the VLA model into a robust and versatile general-purpose solution. One group of studies focuses on searching for better approaches to simulate the action space. For example, instead of discretizing robot actions into language tokens, these studies treat raw continuous actions as the predictive target. Therefore, these works \citep{li_vision-language_2023,li_cogact_2024,liu2025rdtb,li_towards_2024,wen_tinyvla_2025,wen_diffusion-vla_2025} employ a separate policy head, using an RNN, Transformer\citep{vaswani_attention_2017}, diffusion network\citep{chi_diffusion_2023}, or the flow matching\citep{black__0_2024} technique, to enhance the model performance. Another group of studies\citep{zhen_3d-vla_2024,barcellona2025dream} focuses on promoting the VLA by incorporating augmented robot observations, such as improving representations, utilizing historical information, or grounding robot actions in a world model.

However, most previous studies place an emphasis on directly improving the performance (e.g., success rate) of the VLA, while overlooking the enhancement of the interactive process between humans and robots. Specifically, previous VLA models typically use pure linguistic instructions to guide the robot in performing a task, whereas in daily life, human instructions are not confined to a single form. For example, human guidance could take the form of optical instructions on a whiteboard, a combination of text and images, a one-shot video demonstration, or simply a goal image. To address this issue, we present OE-VLA, a novel approach capable of handling the aforementioned open-ended multimodal human instructions. Accompanied by our method, we also introduce two new benchmarks, OE-CALVIN$_{base}$ and OE-CALVIN$_{hard}$, to ensure reproducibility and comparability. These two benchmarks are built upon the popular CALVIN test suite and feature varying levels of difficulty for open-ended instructions. Experimental results show that the proposed OE-VLA not only matches the performance of traditional VLA models in following language instructions but also demonstrates acceptable performance in tackling diverse free-form human prompts. In conclusion, our contributions are three-fold:
\begin{itemize}
    \item We introduce OE-VLA, a novel VLA model capable of processing diverse open-ended multimodal human instructions through a unified neural architecture.
    \item We propose a general approach for constructing robotic datasets with free-form multimodal instructions by leveraging existing datasets. Meanwhile, a two-stage curriculum learning algorithm is proposed to fine-tune a vision language foundation model, capable of handling interleaved modalities, into a VLA model for open-ended instructions.
    \item We introduce two robust benchmarks with diverse open-ended instructions, based on the CALVIN suite, to evaluate our method.
\end{itemize}

\begin{figure}[t]
    \centering
    \includegraphics[width=1.0\textwidth]{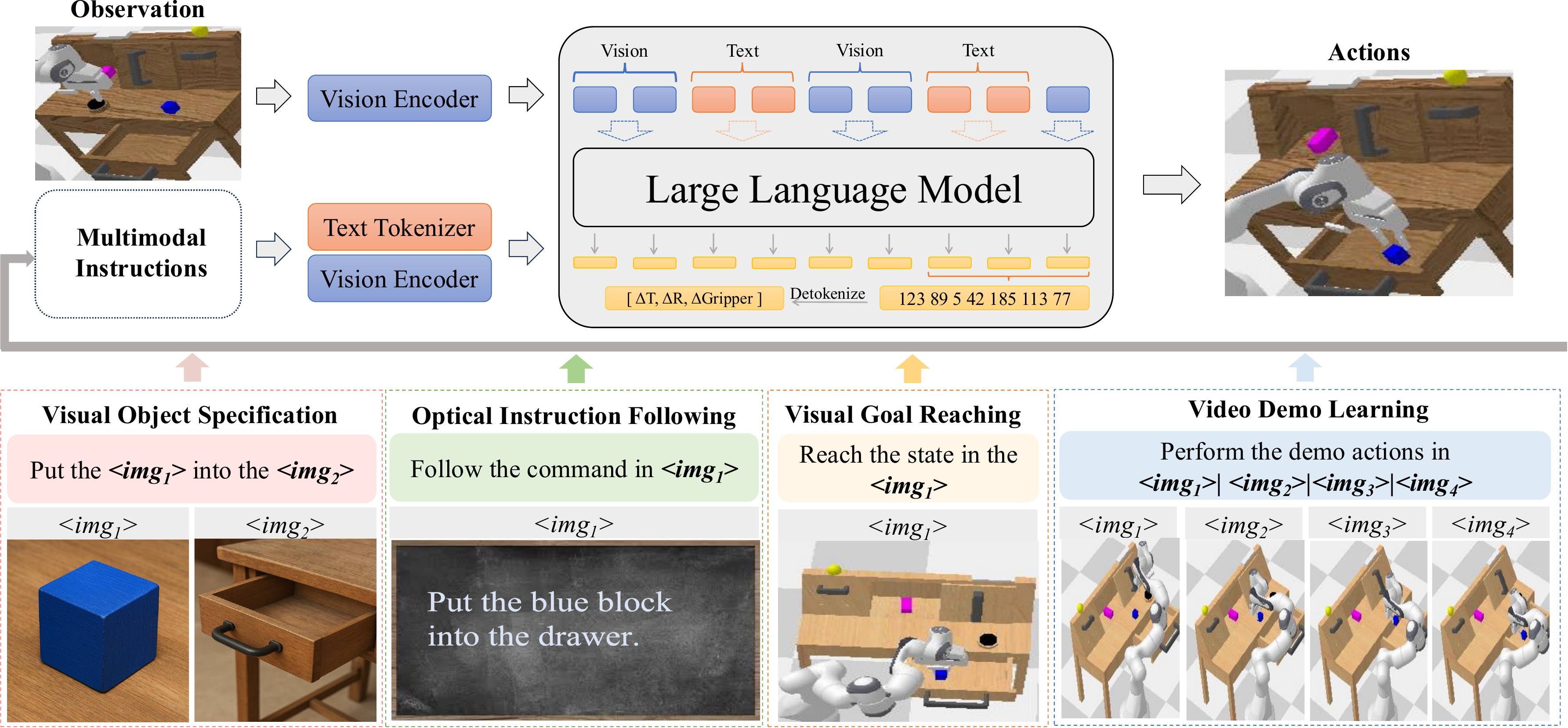}
    \caption{The introduced OE-VLA with four open-ended robot manipulation tasks.}
    \label{fig:fig1}
\end{figure}

\section{Related Work}
\label{sec:2}
Previous studies, such as RT-2\citep{zitkovich_rt-2_2023}, RT-X\citep{collaboration_open_2023}, and OpenVLA\citep{kim_openvla_2024}, use a pretrained VLM as the foundation and fine-tune it into a VLA model with carefully curated robot demonstration datasets. Leveraging the vast knowledge contained within the VLM, these models outperform traditional robot policies, yielding remarkable results. RoboFlamingo\citep{li_vision-language_2023} has made further efforts to incorporate historical visual information as input and predict continuous robot actions, rather than relying on discretized action tokens. Following the remarkable success of diffusion policy\citep{chi_diffusion_2023}, numerous VLA studies, such as CogACT\citep{li_cogact_2024}, $\pi_{0}$\citep{black__0_2024}, and GR00T\citep{nvidia_gr00t_2025}, have integrated it as a separate output head and achieved SOTA performance. Numerous works have also attempted to improve VLA performance from other perspectives. 3D-VLA\citep{zhen_3d-vla_2024} leverages depth information, RT-H\citep{belkhale_rt-h_2024} introduces action hierarchy techniques, and Embodied-COT\citep{zawalski2024robotic} integrates classic chain-of-thought reasoning into the VLA model. Moreover, there is also a line of work\citep{wu_daydreamer_2023,barcellona2025dream} that leverages world models to further enhance performance.

However, previous studies have predominantly concentrated on improving the success rate of robot manipulation using language modality instructions. Few studies have explored equipping VLA models with the capability to handle more open-ended task specifications. Perhaps the study that has the most similar motivation to ours is VIMA\citep{jiang_vima_2023}, which is capable of receiving multimodal prompts for robot control. However, our study differs significantly in many aspects. First, VIMA is an object-centric robot policy model, rather than a traditional VLA model. It requires an object detector to segment each object from the observations and process the combination of object images and textual prompts. Thus, it does not support the pure language-conditioned working pattern as VLA does. Second, VIMA is built on a language model, such as T5, which lacks knowledge about the visual world. Third, the multimodal instructions in VIMA are constrained to originate from the same working space. In contrast, our model is capable of handling instructions that may come from different environments or even from the Internet. We are pleased to note that a concurrent work\citep{fan_interleave-vla_2025} shares a similar motivation to ours; however, our focus is on addressing more open-ended tasks.


\section{Open-Ended Tasks with Multimodal Instructions for VLA}
\label{sec:3}
In addition to the original language-conditioned tasks, this study further investigates the following diverse open-ended tasks for VLA as depicted in Figure \ref{fig:fig1}.

\begin{itemize}
    \item \textbf{Visual object specification (VOS).} In this task, the objects to be manipulated are represented by a snapshot image rather than a language description. The task prompt follows a format like “Given observation <obs>: \emph{grasp the <img$_1$> and put it into <img$_2$>},” where <img$_1$> and <img$_2$> represent the relevant objects, <obs> is the original observation image in VLA.

    \item \textbf{Optical instruction following (OIF).} For tasks in this category, the robot must follow instructions presented in an image, rather than in a textual form. The task prompt is structured as “Given observation <obs>: \emph{follow the command in <img>}.”

    \item \textbf{Visual goal reaching (VGR).} This task requests the robot to identify the differences between the observed state and the goal state, then take appropriate actions to achieve the goal state. The task prompt is expressed as “Given observation <obs>: \emph{reach the goal state in <img>}.”

    \item \textbf{Video demo learning (VDL).} In this task, the robot is provided with a single video demonstration consisting of about four frames that show what the robot is required to do. The robot is expected to infer what happens in the video and take the appropriate actions. The whole prompt is "Given observation <obs>: \emph{perform the demonstrated actions in <img$_1$> <img$_2$> <img$_3$> <img$_4$>}," where <img$_i$> represents the video frame at step $i$.
\end{itemize}

\section{Method}
\label{sec:4}

\subsection{Overall Architecture}

\begin{figure}[htbp]
    \centering
    \includegraphics[width=0.5\textwidth]{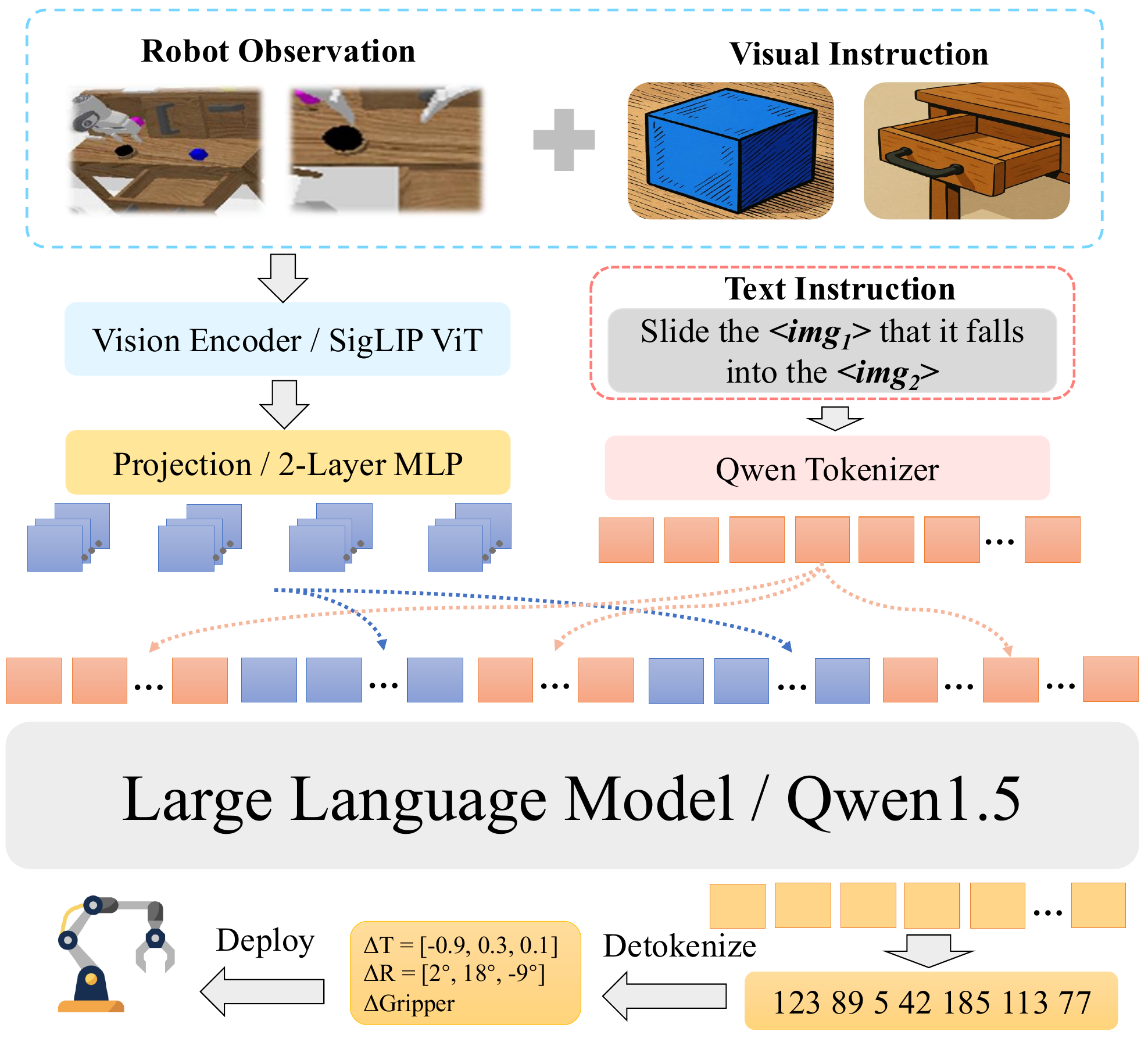}
    \caption{The comprehensive architecture of our OE-VLA model.}
    \label{fig:fig2}
\end{figure}

Figure \ref{fig:fig2} illustrates the overall architecture of our model, which consists mainly of three components: a vision encoder, an MLP projector, and an LLM backbone. Since our work focuses on endowing the VLA with the ability to handle diverse open-ended tasks, we do not incorporate a separate policy head, such as the diffusion policy, to pursue further performance enhancement. Therefore, we adopt classical discretized action tokens as the predictive target, which can be directly generated by the LLM backbone. Specifically, we select LLaVA-Next-Interleave\citep{li_llava-next-interleave_2024} as the foundation model due to its capability to handle free-form multi-image input and its outstanding performance on relevant benchmarks.

\paragraph{Vision encoder} The foundation model uses the SigLIP-400M\citep{zhai_sigmoid_2023} Vision Transformer (ViT)\citep{dosovitskiy2021an}, a CLIP model trained with a more effective sigmoid loss function. This encoder receives an image with a resolution of 384×384 and splits it into patches using a patch size of 14. In particular, our visual information comes from two sources. One source is for the observation side, which includes a static view of the workspace and a dynamic view from the wrist camera. The other source consists of descriptive images from the instruction. We apply this vision encoder to all images, obtaining two types of tokens as follows:
\begin{align}
        T^{obs} &= SigLip(I^{obs}) \\
        T^{img}_{1}, T^{img}_{2}, ..., T^{img}_{i} &= SigLip(I^{img}_{1}, I^{img}_{2}, ..., I^{img}_{i})
\end{align}

\paragraph{LLM backbone} The LLM backbone used in this foundation model is Qwen-1.5\citep{bai_qwen_2023}, which supports a maximum context length of 32k tokens. Any textual fragments $L_{j}$ in the task prompt are tokenized using the Qwen tokenizer as below:
\begin{align}
    T^{lang}_{1}, T^{lang}_{2}, ..., T^{lang}_{j} = Tokenizer(L_{1}, L_{2}, ..., L_{j})
\end{align}

A two-layer MLP is used to map the visual tokens from the vision encoder to the same hidden space as the language tokens. We concatenate all of these tokens while maintaining their original positions to form the final sequence of tokens that is then fed into the LLM.
\begin{align}
        T^{final} = Concat(T^{obs}, T^{lang}_{1}, T^{img}_{1}, T^{lang}_{2}, T^{img}_{2}, ...)
\end{align}

\paragraph{Action Tokenizer} To enable the foundation model to produce action tokens, the continuous robot actions are discretized into 256 bins. These bins are denoted by reusing the least frequently used language tokens in the Qwen vocabulary. The entire model then constructs a conditional probability distribution for the robot actions as follows.
\begin{align}
    P_{\phi}(A_{1},A_{2},..., A_{n}|T^{final}) = \prod_{i=1}^{n}P_{\phi}(A_i|T^{final},A_{1:i-1})
\end{align}
Using the maximum log-likelihood loss, the model parameters $\phi$ can be determined after training. We use a simple five-step action chunk as the predictive target.

\subsection{Training Data Construction}
\begin{figure}[htbp]
    \centering
    \includegraphics[width=0.8\textwidth]{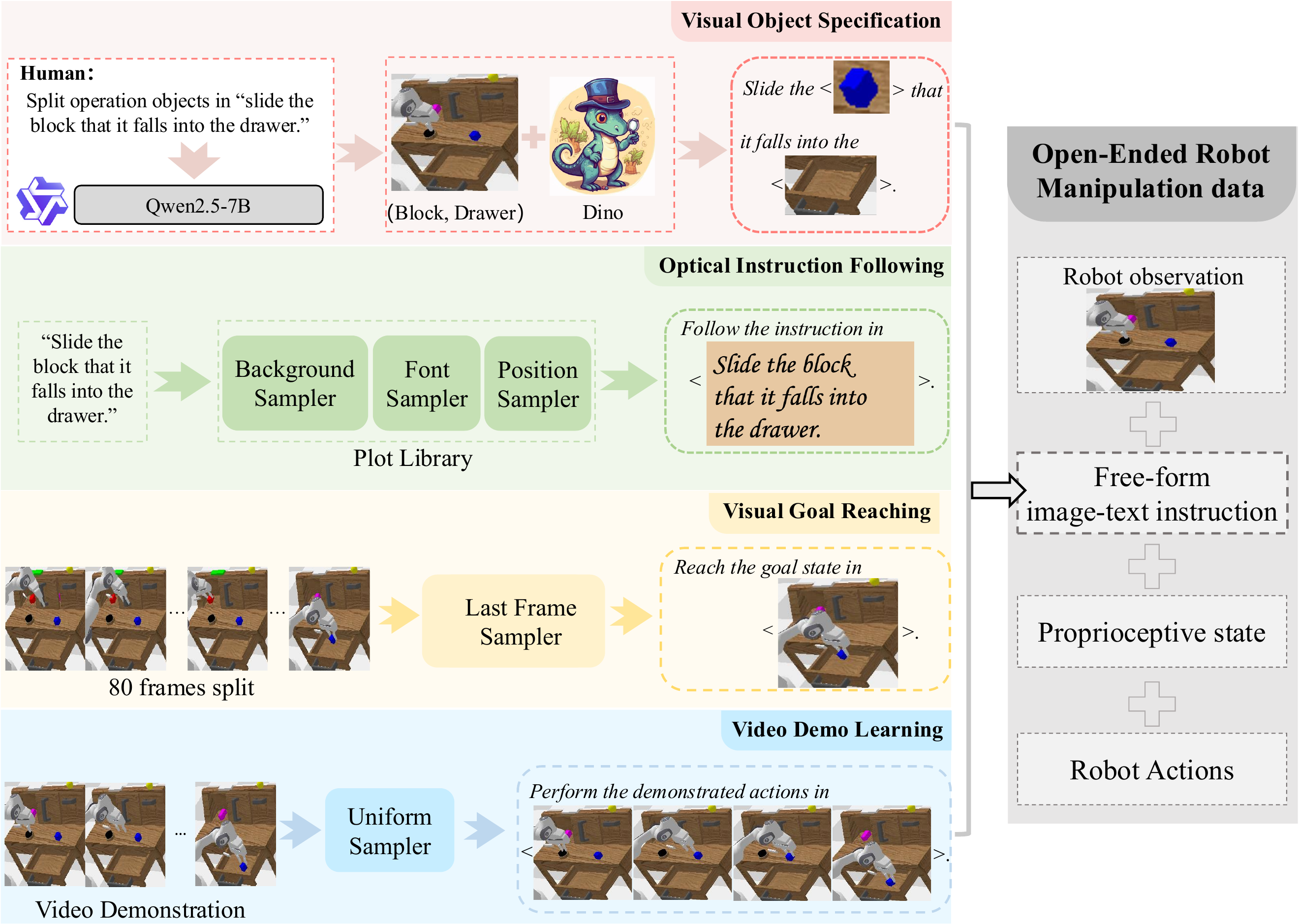}  
    \caption{The data construction method for transforming traditional dataset into the target dataset with open-ended instructions.}
    \label{fig:fig3}
\end{figure}

We propose a simple, yet effective method for transforming any existing robot manipulation dataset with language annotations into a dataset containing open-ended, multimodal instructions. We randomly selected approximately the same number of training samples from the original data for five subsets. One subset is reserved without modification to preserve the original ability of the model to process linguistic instructions. The remaining subsets are transformed into their corresponding open-ended forms as illustrated in Figure \ref{fig:fig3}. 
To construct the visual object specification data, we first utilize an open-source VLM to identify and summarize all potential objects mentioned in the language annotations. Subsequently, an open-vocabulary object detection model is used to detect and localize these objects in the training samples. The images of these objects are later cropped and stored in a database, with additional manual verification. During training, whenever a specific object annotation is encountered, it is replaced with a corresponding image randomly drawn from the pool of candidates. To construct data for optical instruction following, we use Python’s plotting libraries to render textual instructions onto figures with varying fonts, sizes, colors, backgrounds, and positions. To construct the video demonstration data, we replace the linguistic annotation with a sequence of image frames uniformly sampled throughout the entire episode. Finally, for the visual goal-reaching data, we do not construct training samples at the episode level. Instead, we extract fixed-length segments of 80 consecutive image frames from the raw data to serve as training samples. The last frame of each segment is designated as the goal image.

\subsection{Training Pipeline}
We propose a two-stage curriculum learning strategy to endow the OE-VLA model with the capability to handle open-ended instructions.

\paragraph{Stage 1: Multi-Image Grounding}
This training stage is designed to further enhance the foundation model's ability to accurately perceive spatial relationships between objects. Although LLaVA-Interleave-next has been fine-tuned on an interleaved multi-image dataset, its associated tasks are primarily centered on visual storytelling, video captioning, and difference identification. We hypothesize that there is still a gap between these tasks and robot manipulation. Intuitively, spatial relationships among objects play a pivotal role in robotic manipulation. Therefore, we introduce this multi-image grounding stage to help bridge the gap. During this training stage, we finetune our foundation model using the MGrounding dataset introduced by\citep{li_migician_2025}. MGrounding is a compilation of various grounding datasets that include object tracking, referencing grounding, group grounding, and more. It also includes a new free-form multi-image grounding dataset that contains samples such as “\emph{Please find the same person shown in <img$_{1}$> and locate the person in <img$_{2}$>}.” We employ a customized recipe to reduce training time and avoid corrupted or unusable files. Further details are provided in the Appendix.

\paragraph{Stage 2: Open-ended Instruction Tuning}
After fine-tuning the foundation model in Stage 1, we expect our model to demonstrate more accurate environmental perception and be better prepared to perform embodied tasks. Subsequently, we train the model using our constructed robot manipulation datasets with open-ended task specifications. As described in Section \ref{sec:3}, all training samples for these various tasks can be formatted uniformly as ((<obs>, (text$_{1}$, <img$_{1}$>, text$_{2}$, <img$_{2}$>)), <act>), where <obs> refers to the robot observation captured by the camera, (text$_{1}$, <img$_{1}$>, text$_{2}$, <img$_{2}$>)) represents the multimodal human instruction, and <act> corresponds to the actions. During training, these samples are randomly shuffled. In particular, camera images from both the static and wrist views are concatenated into a single image to reduce the number of tokens to be processed. For the sake of simplicity and clarity, we utilize only the static view for figures presented in the paper.

\section{Experiments}
\label{sec:5}

\subsection{Settings}
\begin{figure}[htbp]
    \centering
    \includegraphics[width=1.0\textwidth]{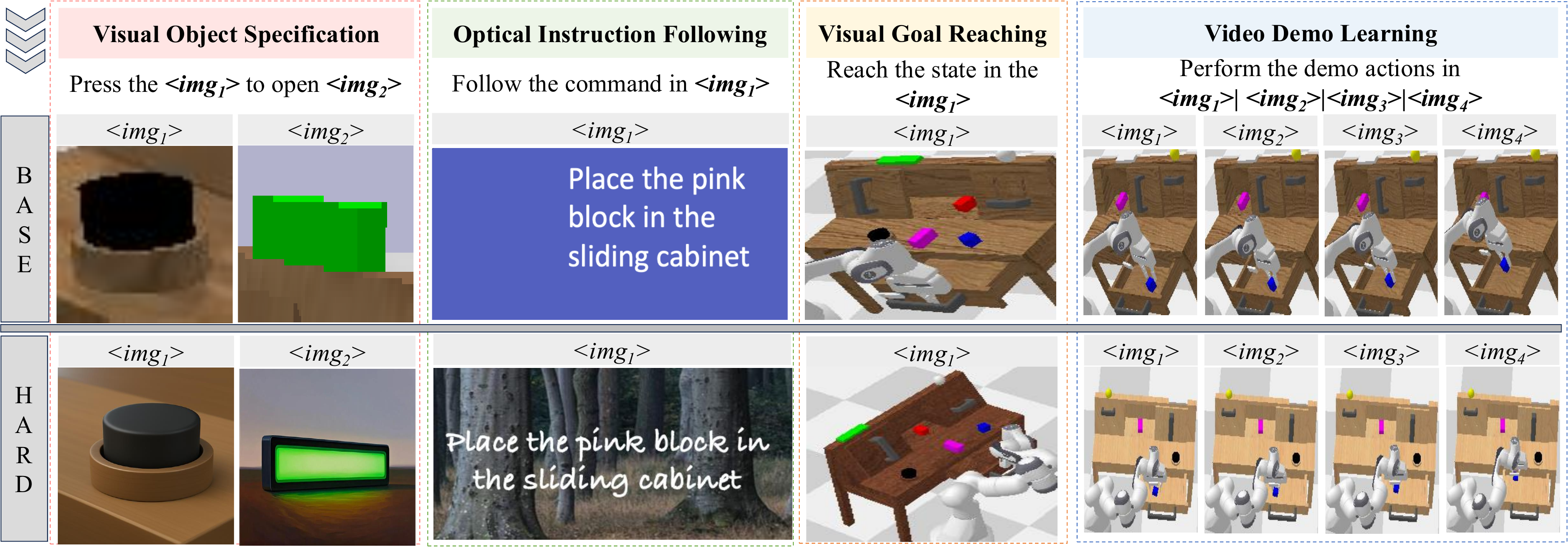} 
    \caption{The introduced two new benchmarks with diverse open-ended task instructions.}
    \label{fig:fig4}
\end{figure}

To ensure reproducibility and facilitate consistent comparisons, our experiments were conducted primarily using the CALVIN\citep{mees_calvin_2022} evaluation suite. CALVIN serves as a robust benchmark for language-conditioned robot policy learning, comprising 1,000 evaluation sequences across 34 tasks to assess model performance. Each evaluation sample consists of a long-horizon task sequence comprising five consecutive subtasks, each accompanied by its corresponding language annotation. In addition, to evaluate our OE-VLA under open-ended instructions, we introduce two new benchmarks of increasing difficulty: OE-CALVIN$_{base}$ and OE-CALVIN$_{hard}$. Each benchmark also contains approximately 1,000 evaluation sequences in total, similar to those in CALVIN. However, all language annotations are replaced with open-ended, multimodal instructions. For the OE-CALVIN$_{base}$ benchmark, the relevant instructions are constructed using object images cropped from raw environmental observations, optical instructions featuring plain backgrounds and regular fonts, and goal images and video demonstrations captured from the same environment without changes in perspective. For the OE-CALVIN$_{hard}$ benchmark, the instructions are constructed using object images sourced from the Web, optical instructions featuring special backgrounds and hand-written fonts, and goal images and videos captured from diverse environments and perspectives. Figure \ref{fig:fig4} presents representative task samples from our OE-CALVIN$_{base}$ and OE-CALVIN$_{hard}$ benchmarks. Additional details regarding these benchmarks and other experiments are provided in the Appendix.

\subsection{Results}
We conducted experiments on the CALVIN benchmark using two data set-split configurations: ABC$\to$D and ABCD$\to$D. The results for the ABC$\to$D split are mainly reported and discussed, as this configuration poses a greater challenge. Meanwhile, we trained both a 0.5B-parameter model and a 7B-parameter model to validate our method. Since no previous VLA models have been capable of handling such open-ended tasks, we introduce strong traditional language-conditioned models as baselines to address the following concerns. \textbf{\emph{1. How does our method perform compared to traditional methods, both using language instructions?}} \textbf{\emph{2. How does our model perform on multimodal open-ended instructions compared to its performance on exclusively linguistic inputs?}} To ensure a fair comparison, we select baseline models that do not incorporate a separate diffusion policy head, since such a head can improve the performance across all models\citep{li_towards_2024}.

\subsubsection{Results with Linguistic Task Specifications}

\begin{table}[htbp]
\caption{Performance of different robot policy models with ABC$\to$D splitting.}
\label{tab:calvin_abc}
\begin{center}
\renewcommand{\arraystretch}{1.0}
\begin{tabular}{l|c|c|ccccc|c}
\toprule
\textbf{Models}     & \textbf{VLA} & \textbf{Base VLM} & \textbf{LH-1}   & \textbf{LH-2}   & \textbf{LH-3}            & \textbf{LH-4}            & \textbf{LH-5}            & \textbf{Len} \\ 
\midrule
MCIL                     &  \ding{55}&  -                        & 30.4\%        & 1.3\%         & 0.2\%         & 0.0\%         & 0.0\%         & 0.31         \\
RT-1                     &  \ding{55}&  -                        & 53.3\%        & 22.2\%        & 9.4\%         & 3.8\%         & 1.3\%         & 0.90         \\
RoboFlam.                &  \ding{51}&  Flamingo-3B                 & 82.4\%        & 61.9\%        & 46.6\%        & 33.1\%        & 23.5\%        & 2.48         \\
OpenVLA                  &  \ding{51}&  Prismatic-7B             & 62.8\%        & 18.3\%         & 5.8\%         & 1.8\%         & 1.0\%         & 0.90        \\
LLaVA-VLA                &  \ding{51}&  LLaVA-7B                 & 83.1\%        & 58.4\%        & 34.7\%        & 23.1\%        & 15.1\%        & 2.14         \\
KosMos Inter.            &  \ding{51}&  KosMos-2B                & 82.4\%        & 68.4\%        & \emph{52.4\%}        & \emph{37.6\%}        & \emph{29.6\%}        & \emph{2.70}         \\
OE-VLA$_{1b}$            &  \ding{51}& Interleave-1B             & \textbf{92.3\%}        & \emph{69.5\%}        & 49.5\%        & 34.1\%        & 24.5\%     & \emph{2.70}         \\
\rowcolor{gray!30}
OE-VLA$_{7b}$          &  \ding{51}& Interleave-7B                & 91.8\%        & \textbf{73.8\%}      & \textbf{56.2\%}    & \textbf{44.2\%}     & \textbf{33.4\%}        & \textbf{2.99}         \\
\bottomrule
\end{tabular}
\end{center}
\end{table}

As demonstrated in Table \ref{tab:calvin_abc}, although our OE-VLA is designed to accommodate a wide range of open-ended multimodal instructions, it still achieves highly competitive performance when evaluated with pure linguistic input. Our OE-VLA$_{1b}$ attains results comparable to those of the KosMos model that does not use a diffusion policy head. The KosMos recipe has previously been shown to serve as the strongest baseline among various VLA models, even compared to models of larger sizes. In addition, we also include the widely used OpenVLA model for comparison. However, the fine-tuned OpenVLA does not outperform other models. We conjecture that this is primarily due to its reliance on a single-view image as input. Therefore, we also introduce LLaVA-VLA, a VLA model that accepts both static view and wrist view inputs and features an architecture similar to that of OpenVLA. For LLaVA-VLA, we apply the same five-step action chunk as OE-VLA. Nevertheless, the performance of this model is still inferior to that of our OE-VLA and the KosMos model. Specifically, our OE-VLA$_{7b}$ achieves the best performance under the CALVIN ABC$\to$D setting, with an average successful sequence length of $2.99$.

\subsubsection{Results with Open-Ended Multimodal Instructions}

\begin{table}[htbp]
\caption{Performance on our introduced OE-CALVIN$_{base}$ (ABC$\to$D) benchmark.}
\label{tab:calvin_oe_base}
\begin{center}
\renewcommand{\arraystretch}{1.0}
\begin{tabular}{l|c|ccccc|c}
\toprule
\textbf{Models}           & \textbf{Task Type}                 & \textbf{LH-1}            & \textbf{LH-2}            & \textbf{LH-3}            & \textbf{LH-4}            & \textbf{LH-5}            & \textbf{Len}             \\ 
\midrule
\multirow{5}{*}{OE-VLA$_{1b}$}    & VOS        & \textbf{94.0\%}                   & \textbf{78.7\%}             & \textbf{56.3\%}                   & \textbf{42.0\%}                   & 30.0\%                   & \textbf{3.01}                     \\
                                  & OIF      & 92.3\%                   & 70.0\%                   & 51.7\%                   & 35.7\%                   & 24.3\%                   & 2.74                     \\
                                  & VGR               & 88.3\%                   & 63.7\%                   & 39.7\%                   & 24.0\%                   & 14.7\%                   & 2.30                     \\
                                  & VDL                & 93.7\%                   & 72.3\%                   & 55.7\%                   & 42.0\%                   & \textbf{31.3\%}                   & 2.95                     \\
\rowcolor{gray!30}                & Avg.                            & - & - & - & - & - & 2.75 \\
\midrule
\multirow{5}{*}{OE-VLA$_{7b}$}    & VOS        & \textbf{95.0\%}            & \textbf{83.3\%}            & 68.7\%                 & 54.3\%                      & 44.7\%                        & 3.46                      \\
                                  & OIF      & 92.3\%              & 81.3\%            & 70.7\%                 & 61.0\%                      & \textbf{52.3\%}                      & 3.58                      \\
                                  & VGR               & 89.7\%                      & 76.0\%          & 64.0\%                      & 51.7\%                      & 44.3\%                      & 3.26                      \\
                                  & VDL                & 93.3\%               & 80.7\%        & \textbf{72.0\%}               & \textbf{61.7\%}                      & 51.7\%                      & \textbf{3.60}                      \\
\rowcolor{gray!30}                & Avg.                            & - & - & -  & - & - & 3.48                      \\
\bottomrule
\end{tabular}
\end{center}
\end{table}

\begin{table}[htbp]
\caption{Performance on our introduced OE-CALVIN$_{hard}$ (ABC$\to$D) benchmark.}
\label{tab:calvin_oe_hard}
\begin{center}
\renewcommand{\arraystretch}{1.0}
\begin{tabular}{l|c|ccccc|c}
\toprule
\textbf{Models}           & \textbf{Task Type}                 & \textbf{LH-1}            & \textbf{LH-2}            & \textbf{LH-3}            & \textbf{LH-4}            & \textbf{LH-5}            & \textbf{Len}             \\ 
\midrule
\multirow{5}{*}{OE-VLA$_{1b}$} & VOS        & \textbf{85.5\%}                & \textbf{61.8\%}               & \textbf{43.2\%}                   & \textbf{29.7\%}                   &  \textbf{22.6\%}                   & \textbf{2.43}                     \\
                          & OIF      & 73.3\%                   & 48.0\%                   & 27.4\%                   & 16.2\%                   & 8.8\%                   & 1.74                     \\
                          & VGR               & 64.5\%                   & 35.5\%                   & 16.2\%                   & 7.8\%                   & 4.7\%                   & 1.29                     \\
                          & VDL                & 73.6\%                   & 44.9\%                   & 27.4\%                   & 13.9\%                   & 9.5\%                   & 1.70                     \\
\rowcolor{gray!30}                        & Avg.                            & - & - & - & - & - & 1.79 \\
\midrule
\multirow{5}{*}{OE-VLA$_{7b}$}   & VOS        & \textbf{92.9\%}                      & \textbf{77.7\%}                      & \textbf{63.9\%}                      & \textbf{54.4\%}                      & \textbf{43.9\%}                      & \textbf{3.33}                    \\
                          & OIF      & 90.9\%                      & 75.7\%                      & 61.1\%                      & 49.3\%                      & 36.8\%                      & 3.14                      \\
                          & VGR               & 54.7\%                      & 27.7\%                      & 17.2\%                      & 14.5\%                      & 11.1\%                      & 1.25                      \\
                          & VDL                & 88.2\%                      & 71.3\%                      & 55.1\%                      & 46.6\%               & 38.9\%                      & 3.00                      \\
\rowcolor{gray!30}        & Avg.         & - & - & - & -  & -                      & 2.68                      \\
\bottomrule
\end{tabular}
\end{center}
\end{table}

In this section, we evaluate our method on both the OE-CALVIN$_{base}$ and OE-CALVIN$_{hard}$ benchmarks using open-ended instructions. It can be observed from Table \ref{tab:calvin_oe_base} that our method achieves highly promising results. Specifically, the average successful sequence length of OE-VLA$_{1b}$ is 2.75, which is even slightly higher than that of the language-conditioned baseline. We find that OE-VLA demonstrates exceptional performance in visual object specification (VOS) tasks, which are quite common requirements in everyday life. Moreover, the results on the tasks of optical instruction following (OIF) and video demonstration learning (VDL) are also satisfactory. The visual goal reaching (VGR) task may be the most challenging, as the model exhibits a noticeable performance degradation compared to the others. As we scale up our model to OE-VLA$_{7b}$, we observe a substantial improvement in performance, which even exceeds that of prior language-based baselines. Given that the number of image-based instructions is nearly four times greater than that of text-only instructions, this training data configuration allows larger models to achieve significantly improved performance on multimodal tasks. 

Furthermore, our model is evaluated on the more challenging OE-CALVIN$_{hard}$ benchmark. This benchmark requires the model to possess strong generalization capabilities to capture intent from multimodal instructions, which may include web images, images or videos from diverse environments and viewpoints, as well as hand-written commands. The results in Table \ref{tab:calvin_oe_hard} demonstrate that our method is also capable of handling such challenging situations, despite some performance degradation compared to OE-CALVIN$_{base}$. It is notable that our model continues to perform competitively on visual object specification (VOS) tasks. In addition, OE-VLA$_{7b}$ further mitigates performance degradation on this benchmark and demonstrates performance comparable to that of traditional language-conditioned VLAs. Nevertheless, visual goal reaching (VGR) remains difficult due to the limited information contained in a single image. These results demonstrate the potential of using free-form multimodal commands to guide a robot in performing tasks.

\subsubsection{Ablation Studies for Training Pipeline}
\begin{figure}[htbp]
    \centering
    \subfloat[VOS]{
    \begin{minipage}{0.23\textwidth}
        \includegraphics[width=1\linewidth]{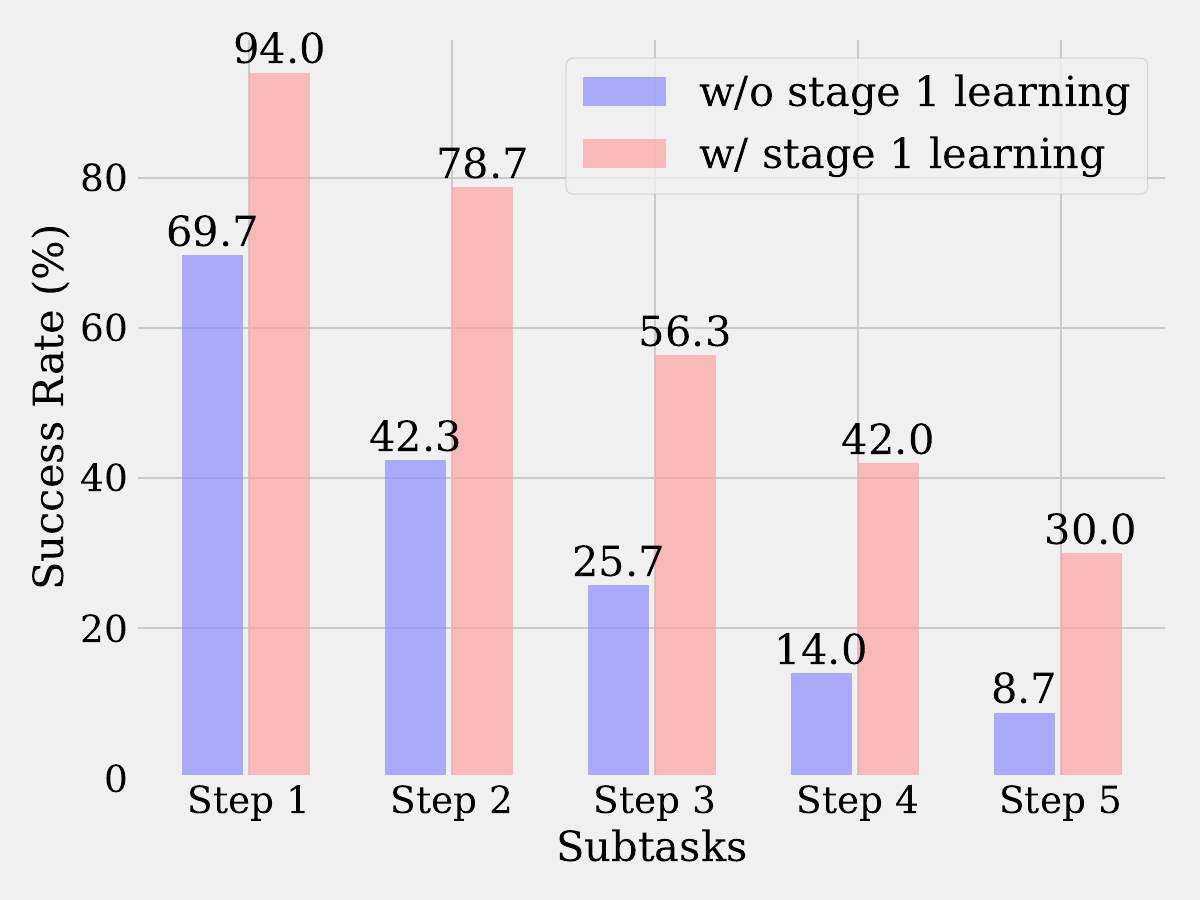}
        \includegraphics[width=1\linewidth]{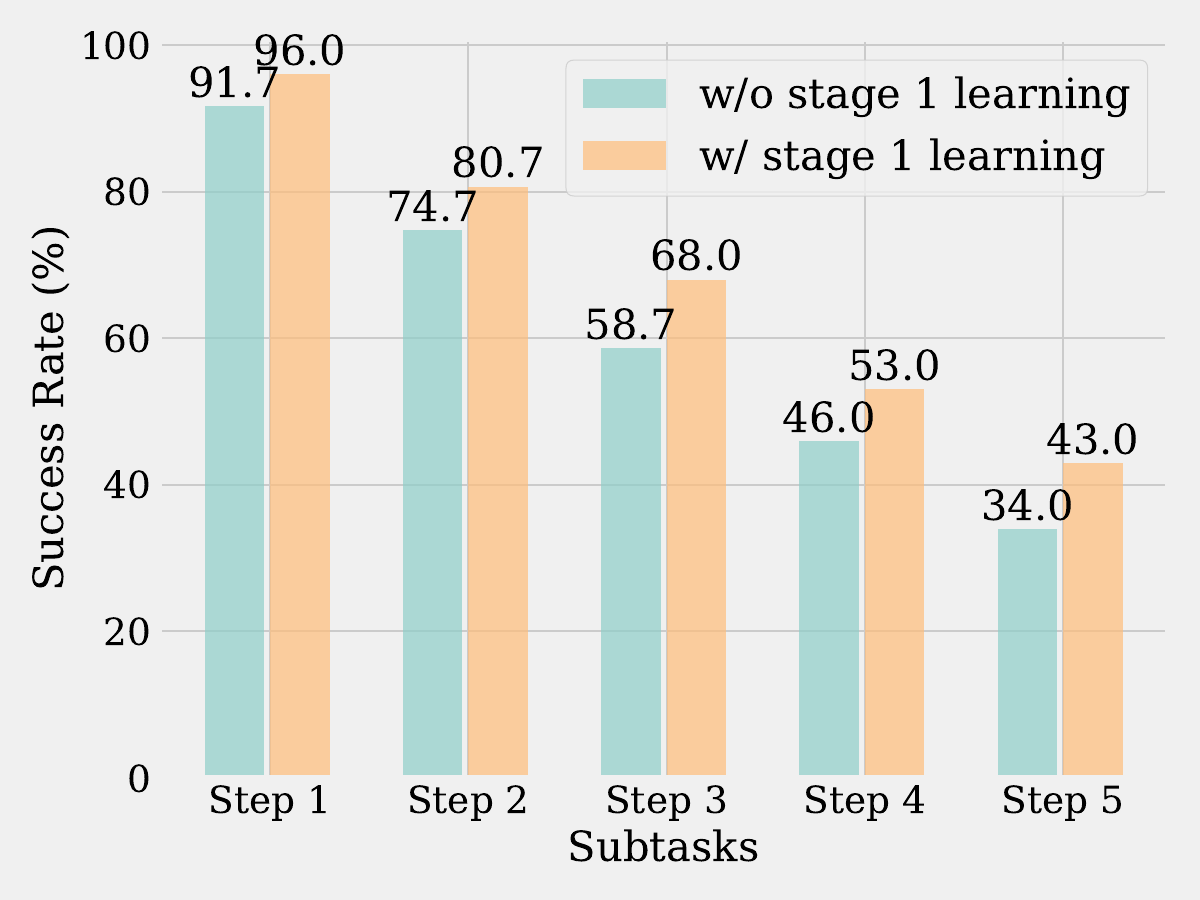}
    \end{minipage}}
    \subfloat[OIF]{
    \begin{minipage}{0.23\textwidth}
        \includegraphics[width=1\linewidth]{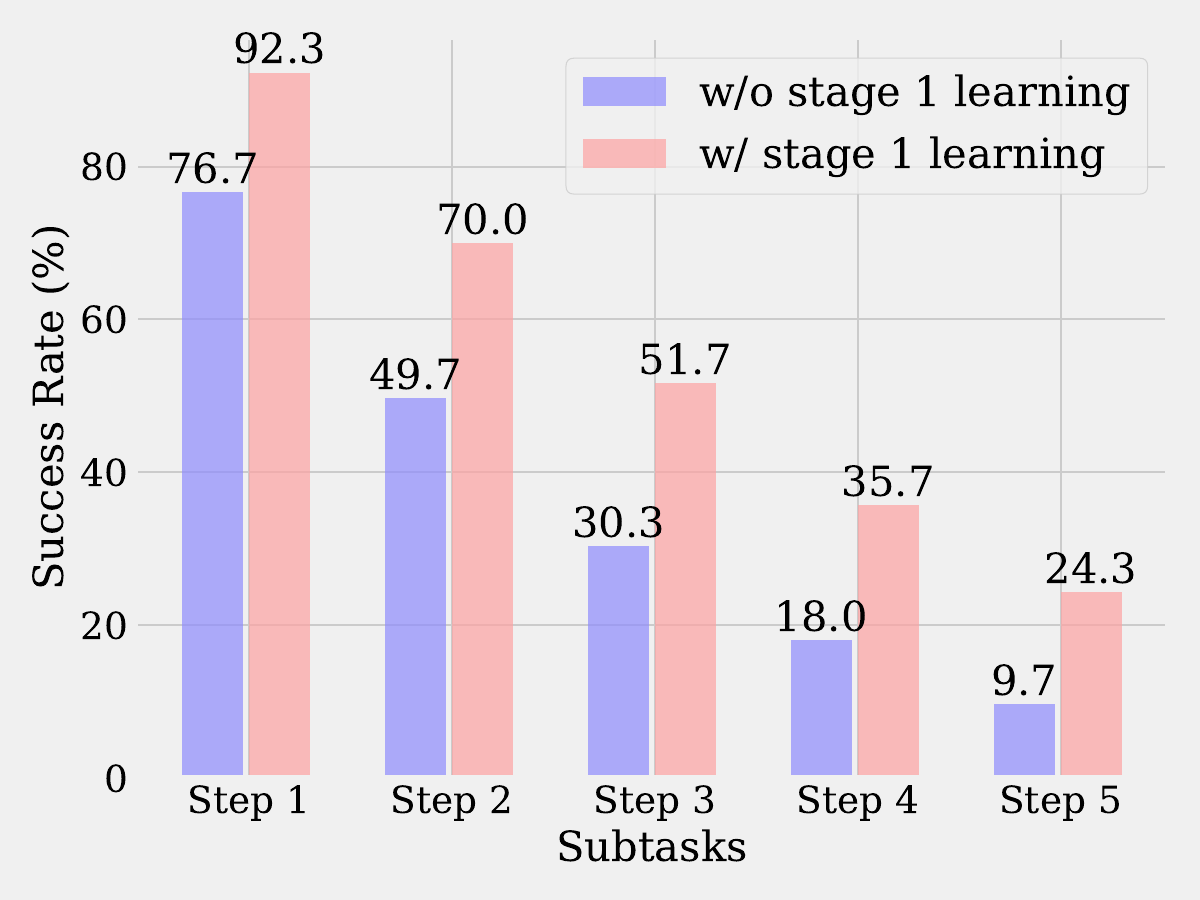}
        \includegraphics[width=1\linewidth]{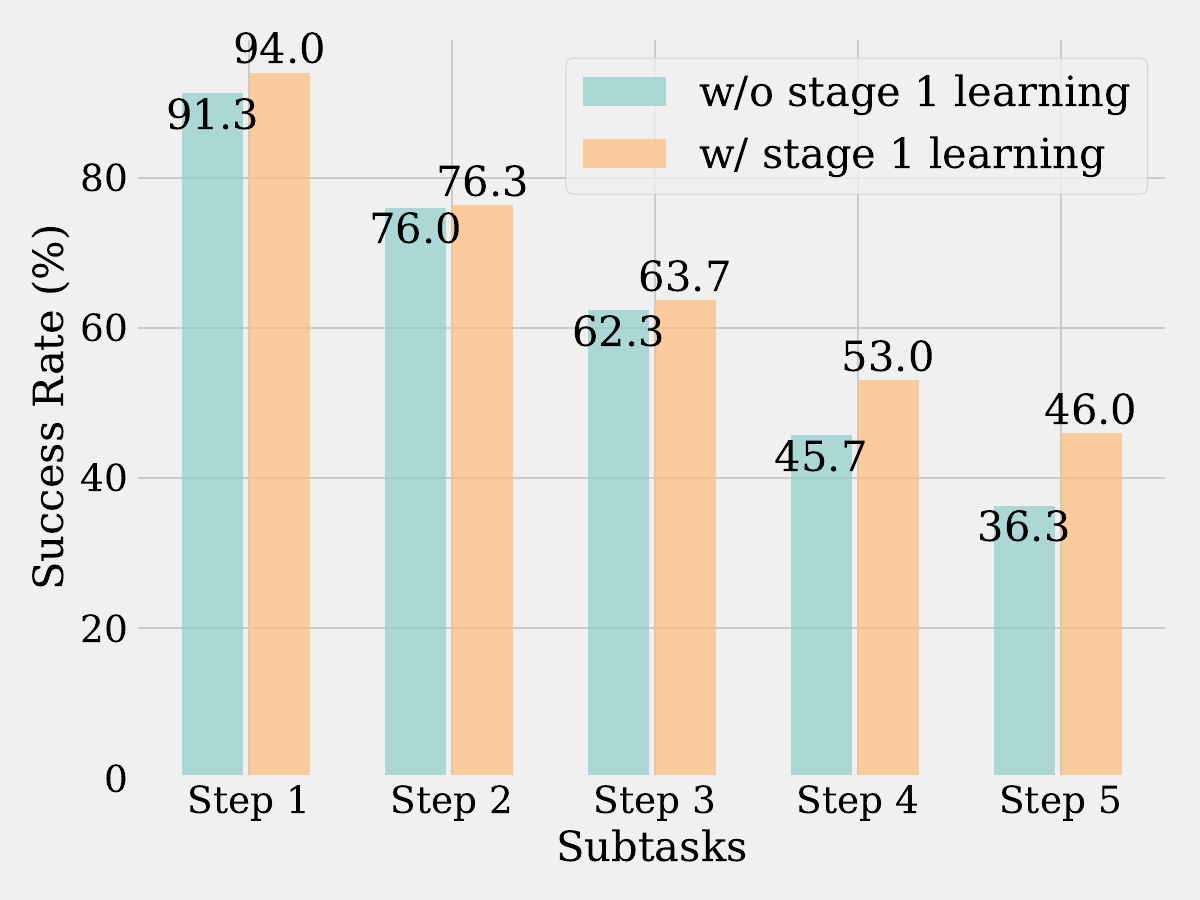}
    \end{minipage}}
    \subfloat[VGR]{
    \begin{minipage}{0.23\textwidth}
        \includegraphics[width=1\linewidth]{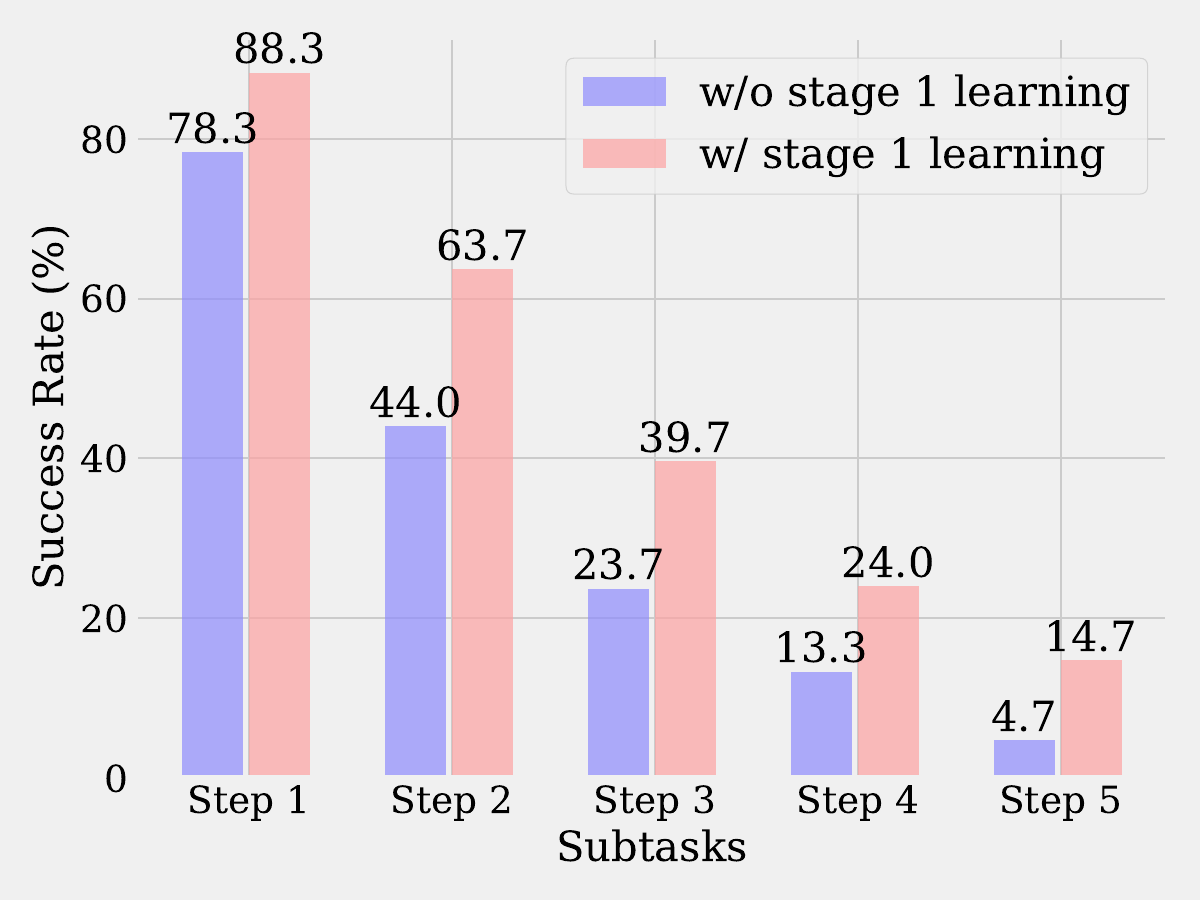}
        \includegraphics[width=1\linewidth]{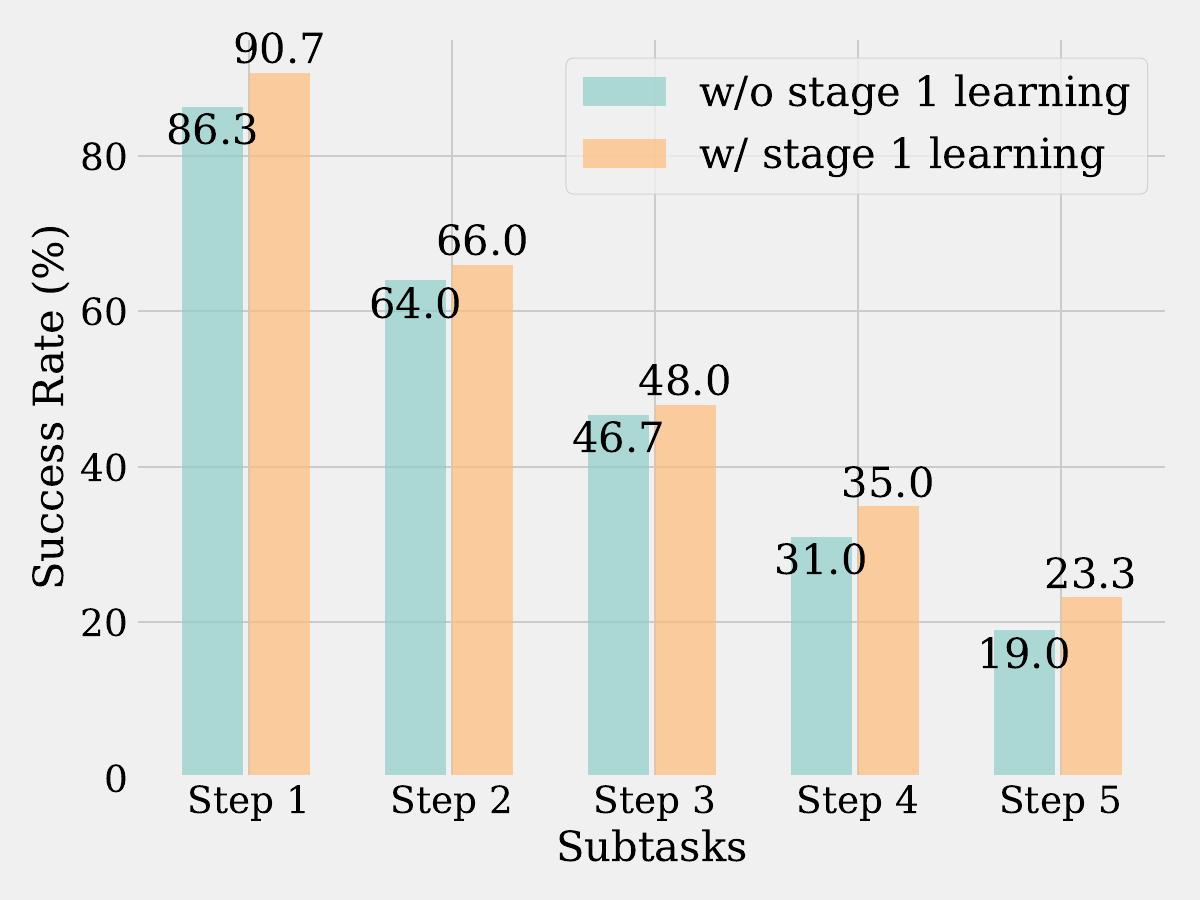}
    \end{minipage}}
    \subfloat[VDL]{
    \begin{minipage}{0.23\textwidth}
        \includegraphics[width=1\linewidth]{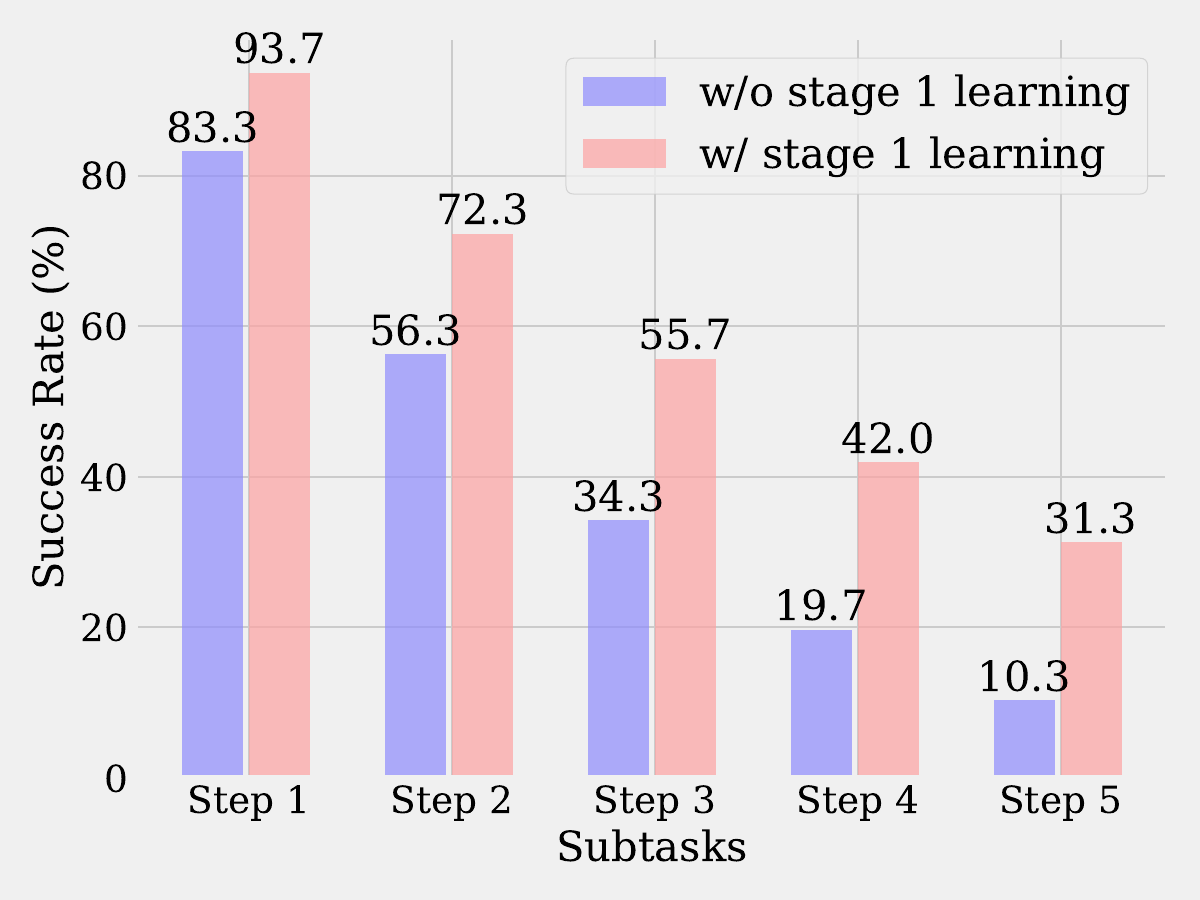}
        \includegraphics[width=1\linewidth]{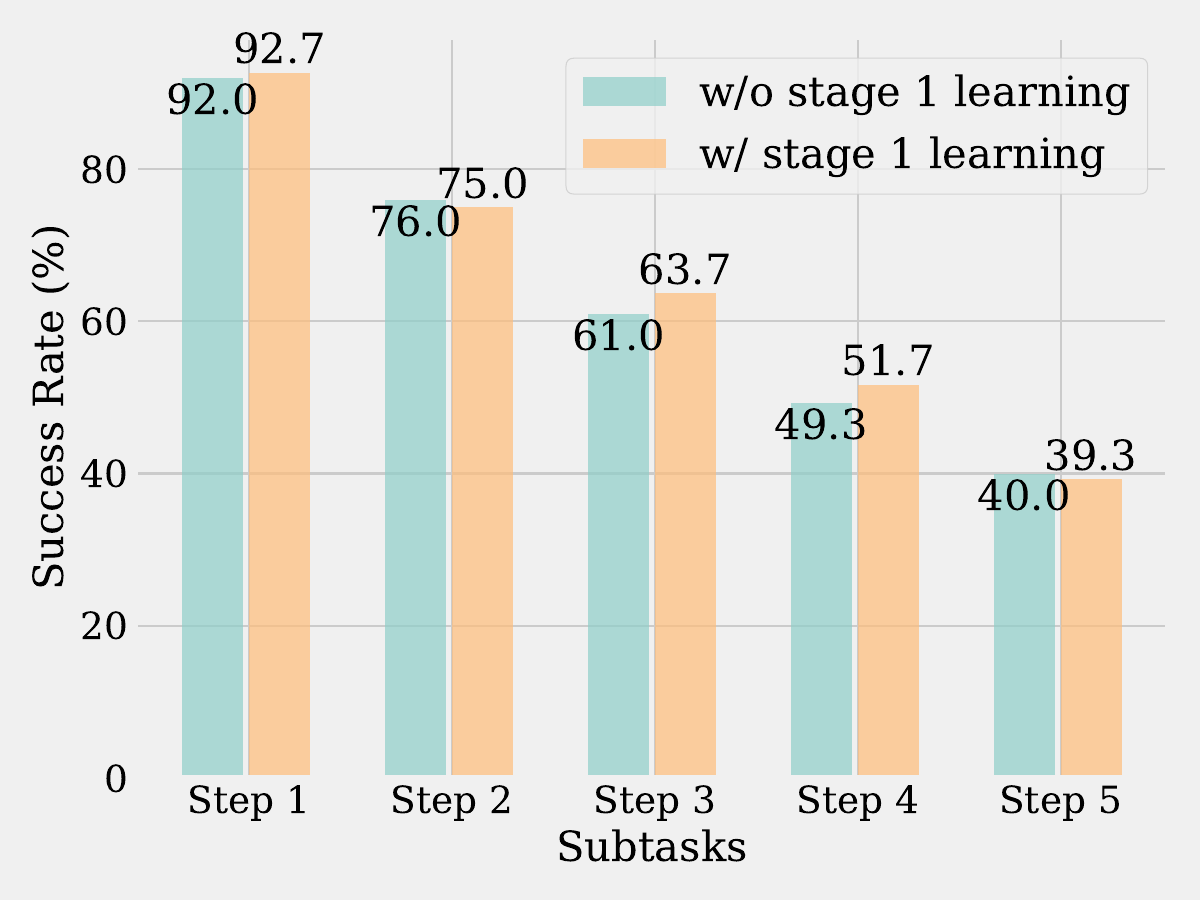}
    \end{minipage}}
    \caption{Performance of our method on the OE-CALVIN$_{base}$ benchmark. The upper row presents the results for the ABC$\to$D split, while the lower row presents the results for the ABCD$\to$D split.}
    \label{fig:ablation_abc_base}
\end{figure}

\begin{figure}[htbp]
    \centering
    \subfloat[VOS]{
    \begin{minipage}{0.23\textwidth}
        \includegraphics[width=1\linewidth]{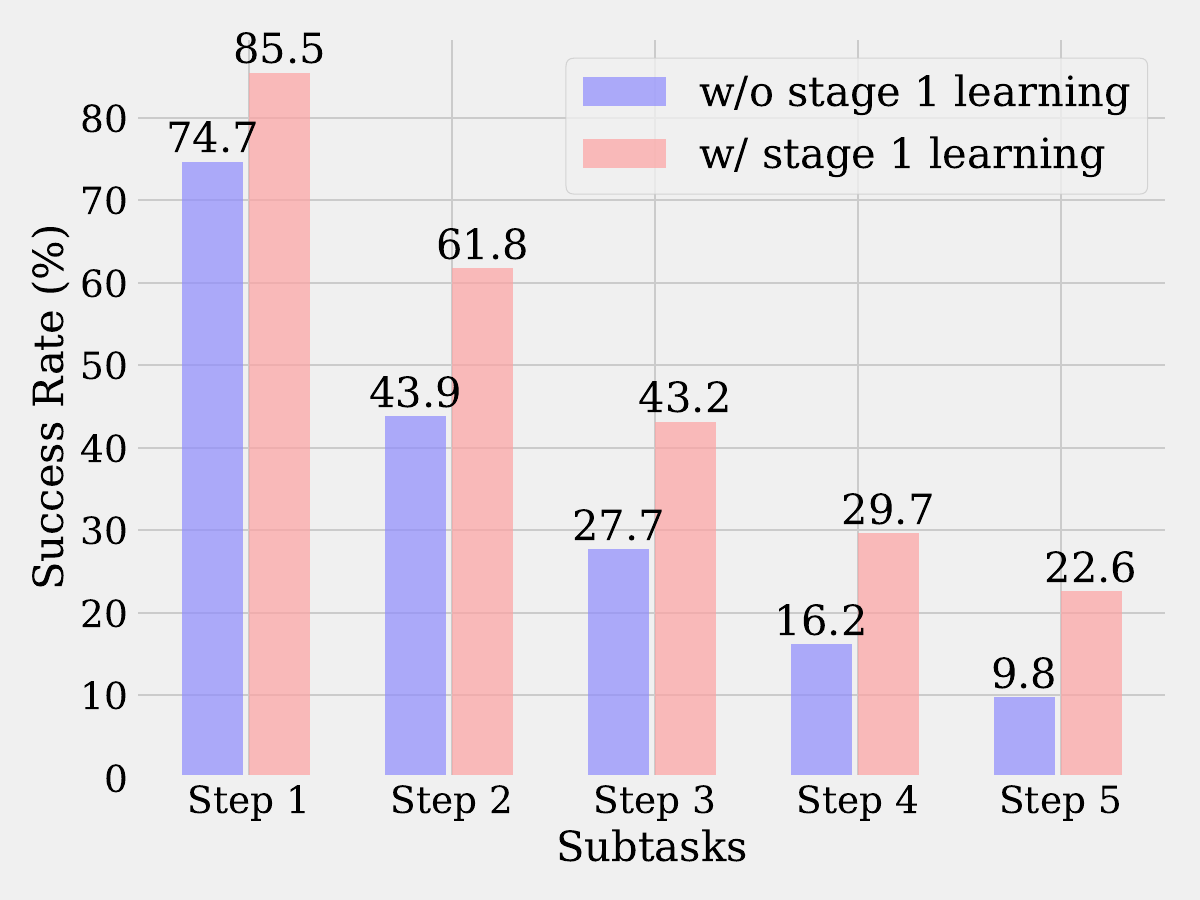}
        \includegraphics[width=1\linewidth]{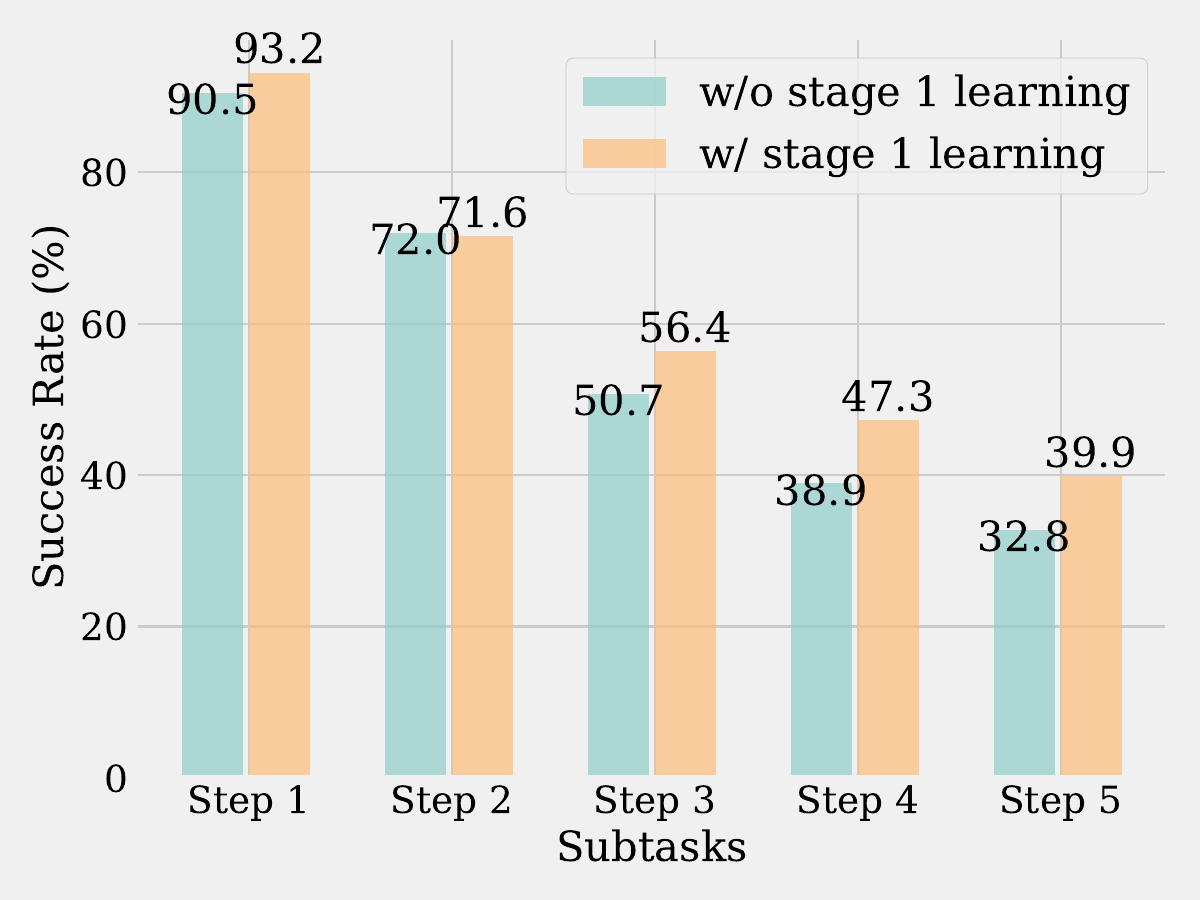}
    \end{minipage}}
    \subfloat[OIF]{
    \begin{minipage}{0.23\textwidth}
        \includegraphics[width=1\linewidth]{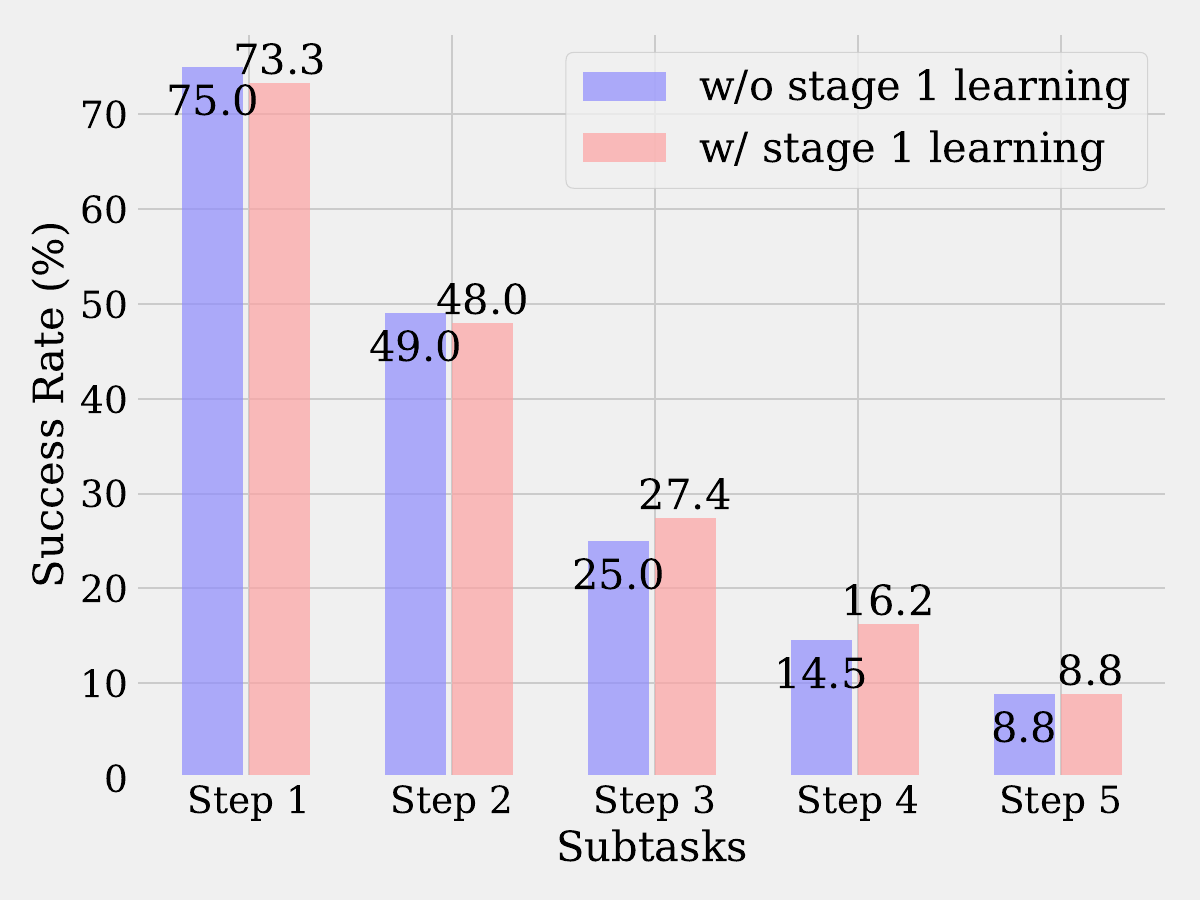}
        \includegraphics[width=1\linewidth]{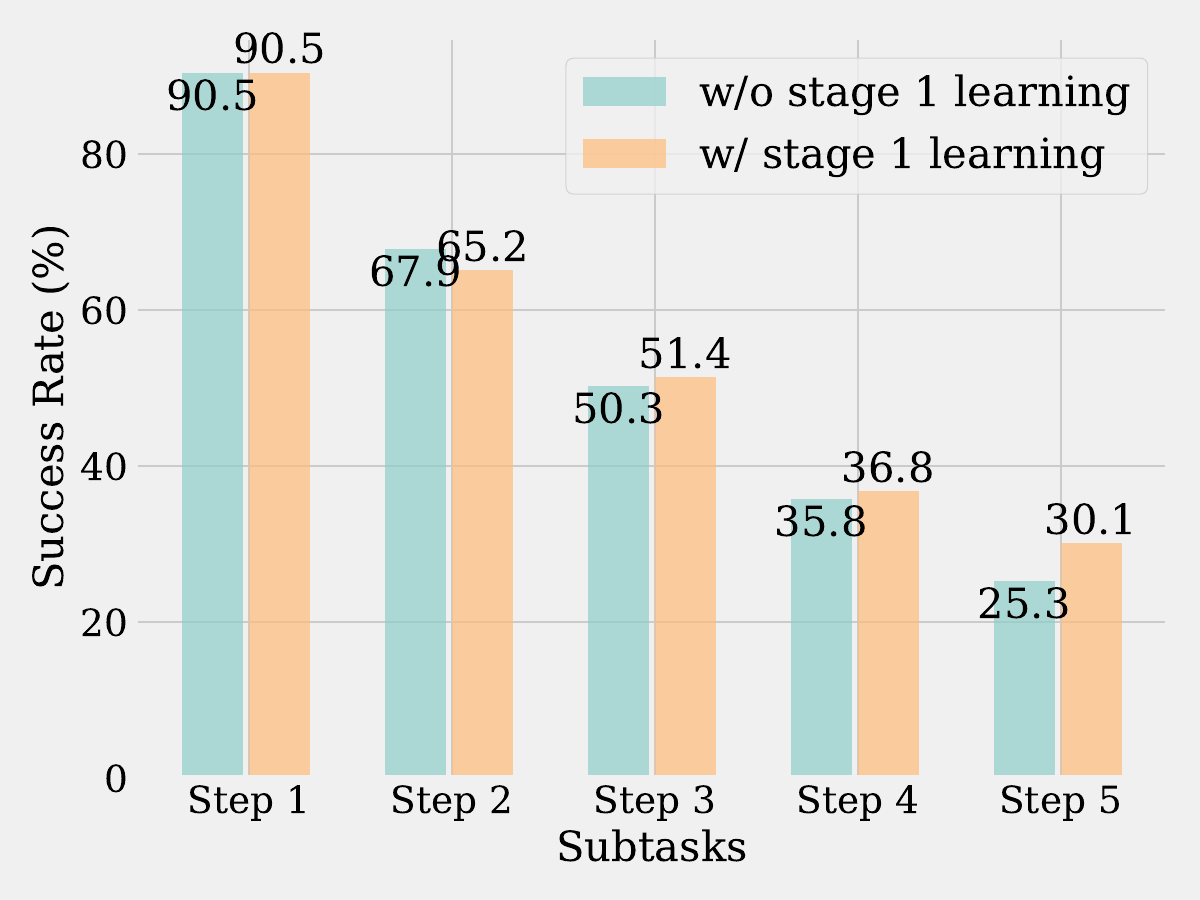}
    \end{minipage}}
    \subfloat[VGR]{
    \begin{minipage}{0.23\textwidth}
        \includegraphics[width=1\linewidth]{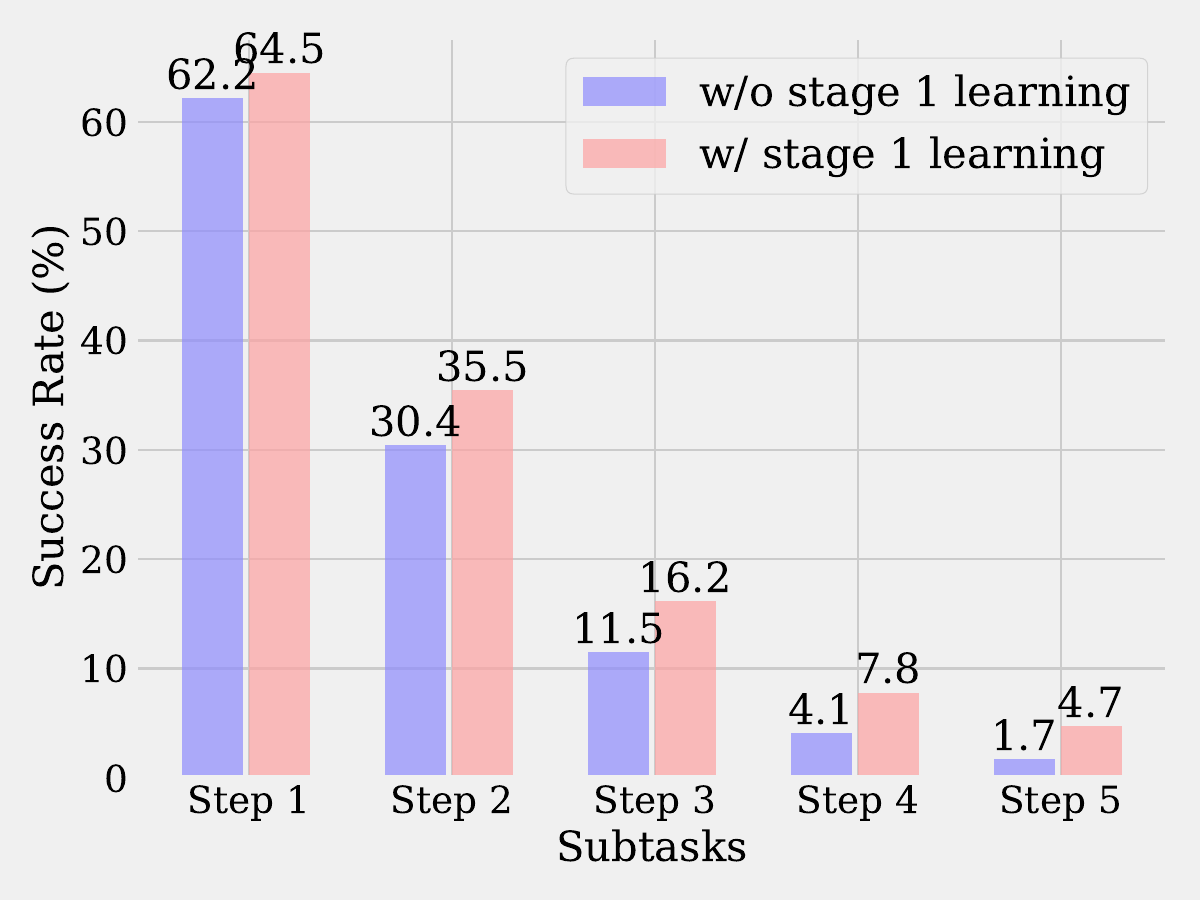}
        \includegraphics[width=1\linewidth]{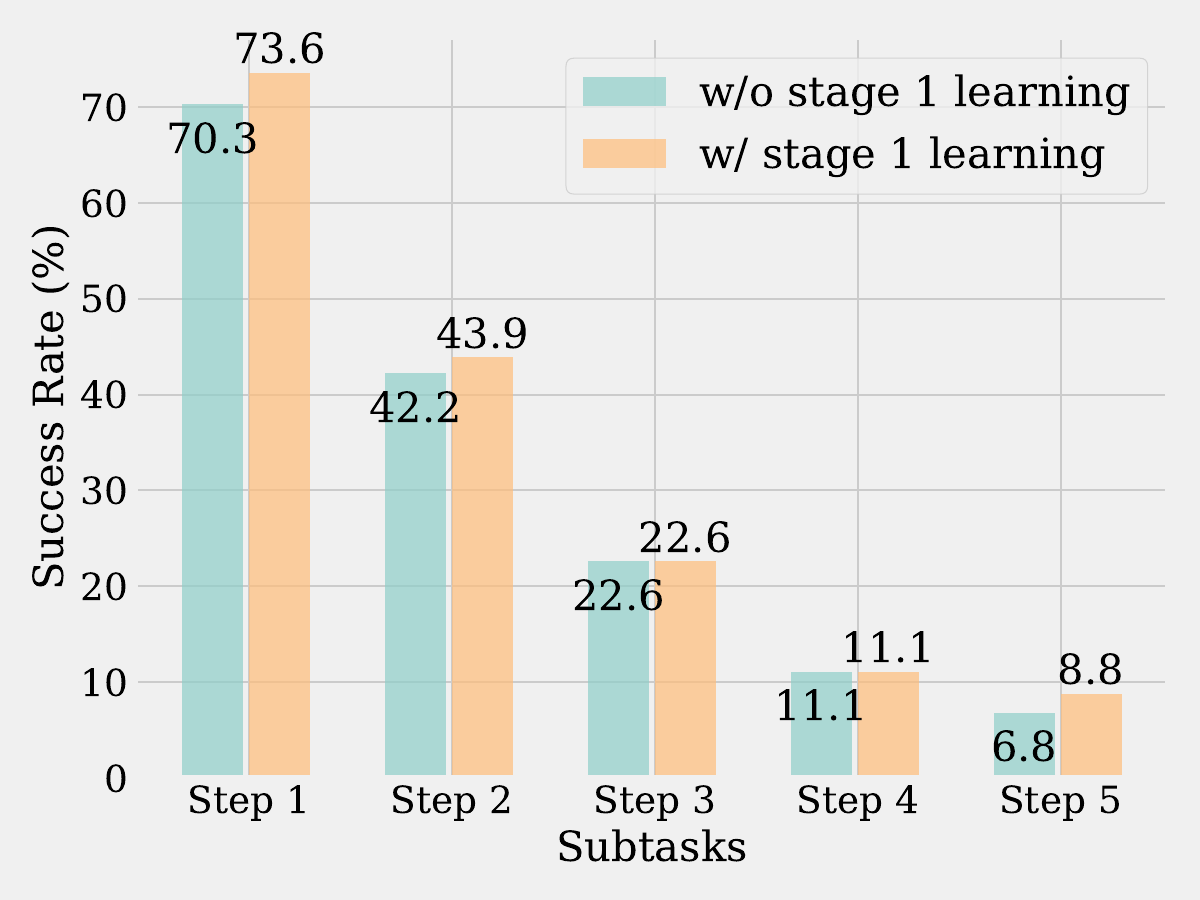}
    \end{minipage}}
    \subfloat[VDL]{
    \begin{minipage}{0.23\textwidth}
        \includegraphics[width=1\linewidth]{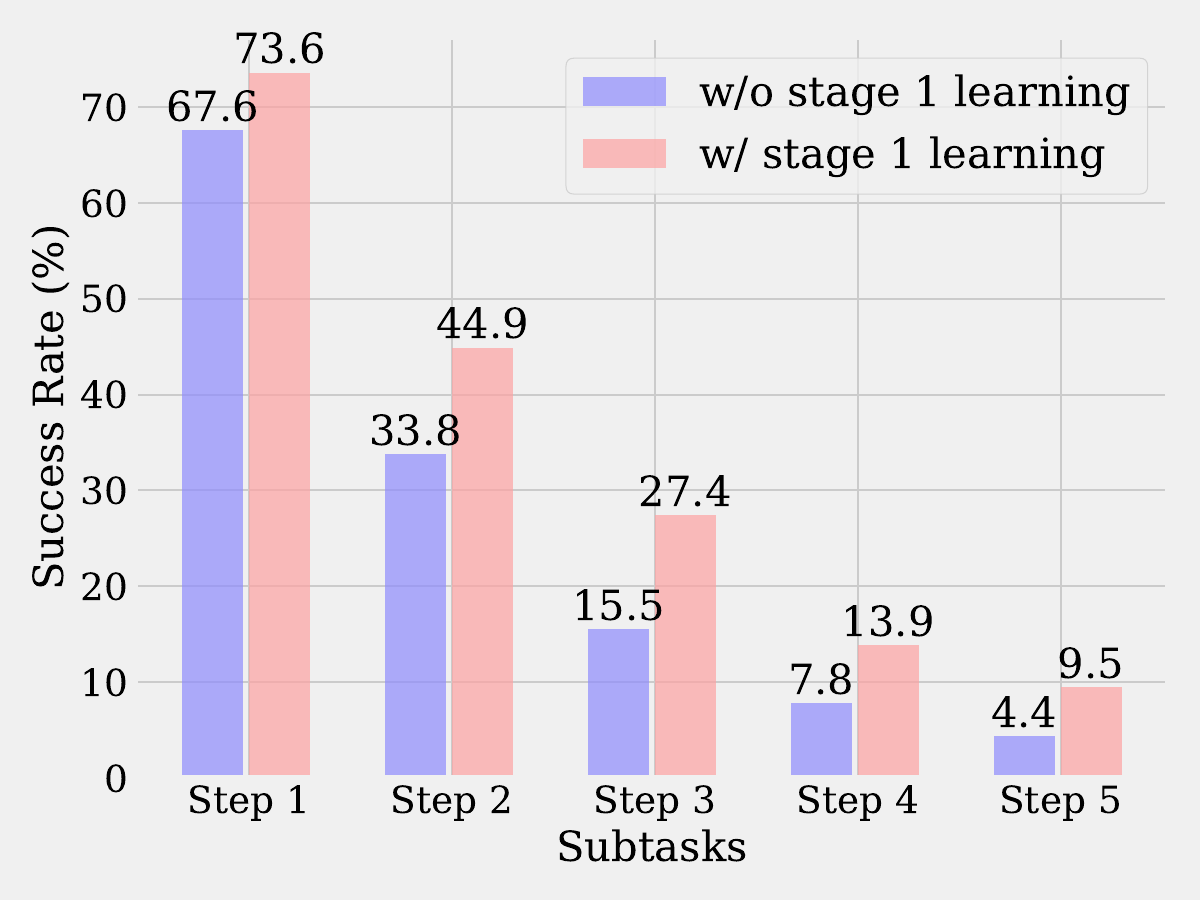}
        \includegraphics[width=1\linewidth]{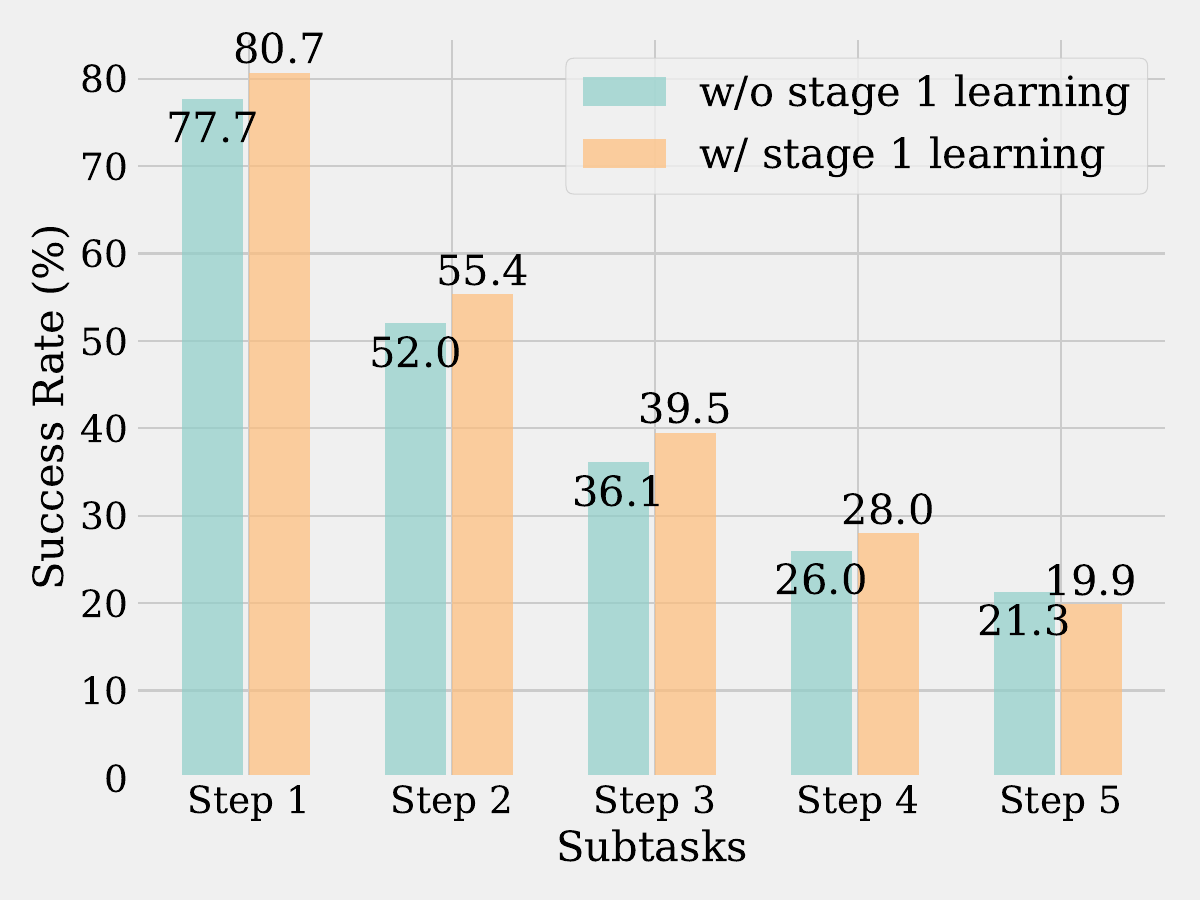}
    \end{minipage}}
    \caption{Performance of our method on the OE-CALVIN$_{hard}$ benchmark. The upper row presents the results for the ABC$\to$D split, while the lower row presents the results for the ABCD$\to$D split.}
    \label{fig:ablation_abc_hard}
\end{figure}

As depicted in Figure \ref{fig:ablation_abc_base}, the two-stage training pipeline substantially improves the performance of our model on the OE-CALVIN$_{base}$ benchmark, particularly for the most challenging ABC$\to$D split. For the ABCD$\to$D split, the improvement is relatively modest. We hypothesize that the model has acquired sufficient knowledge from the training data within the same D environment as the evaluation suite, thus reducing the impact of stage-1 training. However, the proposed two-stage training pipeline still has a positive impact on the overall performance of our model when handling open-ended instructions. A similar phenomenon can also be observed in the results on the OE-CALVIN$_{hard}$ benchmark, as shown in Figure \ref{fig:ablation_abc_hard}. Although the performance gain is less substantial than that on OE-CALVIN$_{base}$, the method is nonetheless effective. Furthermore, our empirical findings suggest that the improvement from stage-1 training for OE-VLA$_{7b}$ is relatively limited, as the foundation model may already possess a strong understanding of spatial relationships.

\subsubsection{Analysis for Training Data Recipe}
As outlined in Section \ref{sec:4}, our multimodal instruction training data are not derived from the original complete dataset, since this would result in a dataset that is five times larger, which significantly needs more time to run experiments. Thus, for each type of task, we randomly select about $40\%$ of the raw data to construct the new training set, which finally results in a dataset twice the original size. Thus, we trained all our models for one epoch for a fair comparison. In particular, the amount of data with multimodal instructions significantly exceeds that of text-only instructions, which may limit the model’s performance on purely linguistic inputs. As shown in Table \ref{tab:calvin_oevla_text}, training models exclusively on textual instructions further improves their performance on linguistic inputs. The results suggest that a better trade-off of the data recipe may be explored in the future.

\begin{table}[htbp]
\caption{Performance of our model trained solely on textual instructions with the ABC→D splitting.}
\label{tab:calvin_oevla_text}
\begin{center}
\renewcommand{\arraystretch}{1.0}
\begin{tabular}{l|ccccc|c}
\toprule
\textbf{Models}      & \textbf{LH-1}   & \textbf{LH-2}   & \textbf{LH-3}            & \textbf{LH-4}            & \textbf{LH-5}            & \textbf{Len} \\ 
\midrule
OE-VLA-text$_{1b}$                                           & 93.2\%        & 76.4\%        & 61.0\%         & 48.7\%         & 39.6\%         & 3.18         \\
OE-VLA-text$_{7b}$                     & 95.1\%        & 83.7\%        & 68.7\%        & 56.1\%        & 45.0\%     & 3.49         \\
\bottomrule
\end{tabular}
\end{center}
\end{table}

\section{Conclusion}
\label{sec:6}
In this work, we presented OE-VLA, a Vision-Language-Action framework designed to handle open-ended multimodal instructions, thereby overcoming the limitations of existing VLA models that rely solely on language input. By enabling the model to interpret various forms of human prompting, including interleaved text and images, handwritten text within images, and demonstration videos, OE-VLA substantially expands the scope of human-robot interaction. Experimental results demonstrate that OE-VLA not only matches the performance of conventional language-driven VLA models but also excels in handling various non-linguistic inputs.

\medskip

\bibliographystyle{unsrtnat}
\bibliography{neurips_2025}

\end{document}